\documentclass[11pt]{article}

\usepackage[encapsulated]{CJK}
\usepackage{ACL2023} % Required for inserting images
\usepackage{times}
\usepackage{latexsym}
\usepackage[T1]{fontenc}
\usepackage[utf8]{inputenc}
\usepackage{microtype}
\usepackage{inconsolata}
\usepackage{graphicx}
\usepackage{array}
\usepackage{listings}
\usepackage{framed}
\usepackage{adjustbox} % 引入adjustbox包

\input{zhwinfonts}

\title{Understanding Literary Texts by LLMs: A Case Study of Ancient Chinese Poetry}

\author{
  Cheng Zhao, Bin Wang, Zhen Wang \\
  \normalsize ByteDance Inc. \\
  \normalsize \texttt{\{zhaocheng.96, wangbin.hd, wangzhen3560\}@bytedance.com}
}

\date{Aug 2024}

\begin{document}

\maketitle

\begin{abstract}

The birth and rapid development of large language models (LLMs) have caused quite a stir in the field of literature. Once considered unattainable, AI's role in literary creation is increasingly becoming a reality. In genres such as poetry, jokes, and short stories, numerous AI tools have emerged, offering refreshing new perspectives. However, it's difficult to further improve the quality of these works. This is primarily because understanding and appreciating a good literary work involves a considerable threshold, such as knowledge of literary theory, aesthetic sensibility, interdisciplinary knowledge. Therefore, authoritative data in this area is quite lacking. Additionally, evaluating literary works is often complex and hard to fully quantify, which directly hinders the further development of AI creation.

To address this issue, this paper attempts to explore the mysteries of literary texts from the perspective of LLMs, using ancient Chinese poetry as an example for experimentation. First, we collected a variety of ancient poems from different sources and had experts annotate a small portion of them. Then, we designed a range of comprehension metrics based on LLMs to evaluate all these poems. Finally, we analyzed the correlations and differences between various poem collections to identify literary patterns. Through our experiments, we observed a series of enlightening phenomena that provide technical support for the future development of high-level literary creation based on LLMs.

\end{abstract}

\section{Introduction}

With the rapid development of LLMs, AI-generated literary texts have made significant progress across various genres. In current practices, the quality of AI-generated works is often judged by human evaluation, which introduces some limitations. Specifically, when the quality of generated works reaches a certain level (e.g., surpassing that of common literature enthusiasts), non-expert evaluations become unreliable, while expert evaluations are scarce and cannot be scaled.

To deepen the understanding of literary works and further improve the quality of AI-generated content, we propose a literature comprehension framework based on LLMs, with a focus on the field of ancient Chinese poetry. This involves three main steps:

\begin{itemize}
    \item \textbf{Anthology Collection}: The collection is a curated assembly of a vast array of historical and AI-generated literary pieces. These works are meticulously sorted into distinct categories, aligned with diverse criteria such as their association with a particular group, the pen of a specific author, or the creative output of a particular AI model. Furthermore, we have engaged esteemed poetry connoisseurs to provide insightful annotations for a select subset of our collection, enriching our compilation with expert insights.
    
    \item \textbf{Metrics Calculation}: We choose a LLM that demonstrates commendable proficiency in Chinese linguistic sphere (denoted as the \textbf{\textit{base}} model). Then we perform fine-tuning and alignment training on historical and annotated samples to obtain a fine-tuned model (denoted as the \textbf{\textit{sft}} model) and an instruct model (denoted as the \textbf{\textit{align}} model). For the entire corpus of poems, we compute an extensive suite of metrics for each model, encompassing yet not limited to: statistical measurements pertaining to tokens, metrics reflective of embeddings and hidden states, metrics indicative of output probabilities, and metrics that mirror relationships (such as attention and covariance matrices), etc. Subsequently, we record these metrics into a database for further analysis.
    
    \item \textbf{Pattern Summarization}: Taking each poetry collection as the research subject, we extract corresponding metrics from the database, and conduct statistical analysis and sampling observation. We select conclusions with high confidence and generalize them into extendable patterns. Accordingly, we optimize model training and generation process based on these patterns.
\end{itemize}

Through this research, we preliminarily establish the viability of employing LLMs to conduct quantitative assessments within the realm of classical Chinese poetry. The valuable patterns we have discerned are poised to markedly elevate the caliber of compositions in future poetic creation endeavors. In this process, we leveraged only a small amount of expert-annotated data to tackle the evaluation challenge in literary texts, demonstrating the high scalability of our approach. We believe that with the support of LLMs, the evaluation of literary works in more genres can be effectively quantified, thereby promoting the prosperous development of AI literary creation.

\section{Related Work}

In recent years, AI's work in literary creation has become increasingly common. As one of the earliest AIGC genres, poetry has garnered much attention with platforms such as ShiSanbai~\footnote{\url{https://www.aichpoem.net/}} and JiuGe~\footnote{\url{https://jiuge.thunlp.org/}}. At the same time, there are also some good papers dedicated to solving the constraints in the poetry generation process and enhancing the aesthetic quality of the generated works, such as MixPoet~\cite{mixpoet:aaai19}, CoPoet~\cite{copoet:emnlp22}, Singer~\cite{singer}, etc.

Witscript~\cite{witscript:corr23} and Witscript 2~\cite{witscript2:corr23} proposed a joke generation system, while Humor Mechanics~\cite{humor_mechanics:corr24} made jokes more appealing by performing multi-step reasoning on LLMs. 

Given the higher difficulty of novel creation, existing work primarily focuses on human-computer interaction and iterative enhancement, with representative works including GROVE~\cite{grove:EMNLP23}, Re3~\cite{re3:emnlp22}, and RELIST~\cite{relist:emnlp22}.

Understanding literary texts through AI tools has long posed a formidable challenge. The study by \cite{eva_literary_llama3:corr24} offered a pragmatic evaluation of Llama3~\cite{llama3}, focusing on its ability to attribute citations within novels. Another investigation~\cite{eva_literary_cor:corr24} delved into the intricacies of coreference within literary texts using LLMs. Additionally, the analysis in \cite{sent_hemingway:jdmdm24} examined the emotional nuances present in Hemingway's writings as a case study. However, these research efforts, while valuable, tend to focus on specific, isolated facets of literature. In contrast, this paper distinguishes itself by pioneering a sweeping, in-depth analysis of literary works facilitated by the capabilities of LLMs, marking a significant departure from the more constrained approaches of previous studies.
 
\section{Methodology}

\subsection{Anthology Collection}

While assessments of literary works inherently involve a degree of subjectivity, several broadly accepted, objective criteria do exist. For example, the poetical oeuvre of a literary luminary such as Su Shi is typically regarded as more accomplished than that of an obscure poet. The poetic styles of the Qing Dynasty are often markedly distinct from those of the Tang Dynasty, which is described as \textit{non-adherence to the grandeur of the Tang}. Similarly, the poetry of the TongGuang style~\footnote{\url{https://baike.baidu.com/item/\%E5\%90\%8C\%E5\%85\%89\%E4\%BD\%93}} are widely acknowledged for their obscurity and complexity, setting it apart from other poetic traditions.

In compiling our anthology, we gathered a batch of poems representing various styles from different historical periods and typical schools. Furthermore, we enlist the expertise of poetry scholars to appraise and annotate a small subset of poems, numbering in the thousands. These annotated samples will serve a dual purpose: they will provide a qualitative assessment and also be utilized for the fine-tuning of LLMs.

We categorize the anthologies into two principal genres: Qilv, a form of traditional Chinese poetry characterized by its stringent metric structure of 8 lines with 7 characters each and Ci, a literary form that flourished from the Song Dynasty, blending poetry and music with a high degree of artistry and expressiveness. The main subcategories within these genres encompass:

\begin{itemize}
    \item \textit{labelled\_good}: A selection of historically acclaimed masterpieces meticulously annotated by experts.
    \item \textit{labelled\_normal}: A compilation of works from history that are deemed average or even subpar by experts.
    \item \textit{historical\_famous}: A collection of works by famous authors from different periods, such as \textit{Three Hundred Tang Poems} and \textit{Three Hundred Song Lyrics}.
    \item \textit{historical\_normal}: Other works by authors whose pieces were labeled as average or subpar by experts, as well as an assortment of random works by lesser-known or anonymous authors.
    \item \textit{ai\_generate}: A showcase of creations from assorted AI models, including but not limited to ShiSanbai, JiuGe, and Singer.
    \item \textit{author\_style}: Exclusive collections highlighting the unique styles of well-recognized authors; for Qilv, this includes Li Shangyin~\footnote{\url{https://baike.baidu.com/item/\%E6\%9D\%8E\%E5\%95\%86\%E9\%9A\%90}}, Huang Tingjian~\footnote{\url{https://baike.baidu.com/item/\%E9\%BB\%84\%E5\%BA\%AD\%E5\%9D\%9A}}, and Chen Sanli~\footnote{\url{https://baike.baidu.com/item/\%E9\%99\%88\%E4\%B8\%89\%E7\%AB\%8B}}; for Ci, it features Su Shi~\footnote{\url{https://baike.baidu.com/item/\%E8\%8B\%8F\%E8\%BD\%BC}}, Zhou Bangyan~\footnote{\url{https://baike.baidu.com/item/\%E5\%91\%A8\%E9\%82\%A6\%E5\%BD\%A6}}, and Gong Zizhen~\footnote{\url{https://baike.baidu.com/item/\%E9\%BE\%9A\%E8\%87\%AA\%E7\%8F\%8D}}, among others.
    \item \textit{group\_style}: Representative works from typical schools; for Qilv, this comprises the TongGuang, JiangXi~\footnote{\url{https://baike.baidu.com/item/\%E6\%B1\%9F\%E8\%A5\%BF\%E8\%AF\%97\%E6\%B4\%BE}}, and XiKun~\footnote{\url{https://baike.baidu.com/item/\%E8\%A5\%BF\%E6\%98\%86\%E4\%BD\%93}}; for Ci, it encompasses the HuaJian~\footnote{\url{https://baike.baidu.com/item/\%E8\%8A\%B1\%E9\%97\%B4\%E8\%AF\%8D\%E6\%B4\%BE}}, ChangZhou~\footnote{\url{https://baike.baidu.com/item/\%E5\%B8\%B8\%E5\%B7\%9E\%E8\%AF\%8D\%E6\%B4\%BE}}, and ZheXi~\footnote{\url{https://baike.baidu.com/item/\%E6\%B5\%99\%E8\%A5\%BF\%E8\%AF\%8D\%E6\%B4\%BE}}, etc.
\end{itemize}

It is worth noting that we ensure a rich diversity in the thematic content of each anthology, thereby mitigating the risk of our research findings being disproportionately swayed by any single thematic element. In the subsequent specially designed theme-related experiments, we unearth several intriguing patterns, the details of which are elaborated in Sec~\ref{sec:discuss}.

\subsection{Metrics Calculation}

We choose the open-source MOSS~\footnote{\url{https://github.com/OpenLMLab/MOSS}} from Fudan University as the \textbf{\textit{base}} model. MOSS is an autoregressive pre-trained model with 16B parameters, boasting impressive comprehension and generation capabilities in Chinese. It encapsulates the broad cognitive grasp of the world by the LLM. 

Later, building upon the \textbf{\textit{base}} model, we conduct LoRA~\cite{lora:iclr22} fine-tuning on a vast collection of historical poems, resulting in the \textbf{\textit{sft}} model. It represents the LLM's mnemonic capacity for historical poetic expressions. 

Subsequently, we advanced our efforts by conducting LoRA alignment training on the annotated samples. This refinement led to the emergence of the \textbf{\textit{align}} model. It represents a significant leap in the LLM's evaluative acumen, manifesting a sophisticated interpretive capability specifically tailored for the poetry in question.

We conduct inferential computations on the entire corpus of poems utilizing our trio of models: \textbf{\textit{base}}, \textbf{\textit{sft}}, and \textbf{\textit{align}}. The input protocol is as follows:

\begin{CJK}{UTF8}{zhsong}
    \textit{以《title》为题写一首七言律诗：content} 
    
    Compose a Qilv titled \textit{title}: \textit{content}
    
    OR
    
    \textit{以《cipai》为词牌名，以《title》为题写一首词：content}
    
    Compose a Ci using the Cipai name~\footnote{\url{https://baike.baidu.com/item/\%E8\%AF\%8D\%E7\%89\%8C\%E5\%90\%8D}} \textit{cipai} and the title \textit{title}: \textit{content}
\end{CJK}

The output comprises the models' inferential analyses on \textit{content}, which has been meticulously stripped of punctuation to ensure accuracy, complemented by data mining techniques. The specific evaluative metrics for each poetic work are delineated below:

\begin{itemize}
    \item \textbf{perplexity}: The measure encompasses both the perplexity of the entire poem and the individual perplexity of each couplet within a Qilv or each section of a Ci, providing a granular assessment of linguistic complexity.
    \item \textbf{out\_entropy}: This item details the information entropy of the model's output probability distribution, alongside the outcomes of the Augmented Dickey-Fuller (ADF) stability test, which evaluates the temporal stationarity of the sequence.
    \item \textbf{abs\_prob}: We collect absolute probability tables for the world, classical Chinese, and Qilv/Ci. After merging the tables, obtain the absolute probability sequence of tokens.
    \item \textbf{prob\_kld}: The Kullback-Leibler (KL) divergence measures the difference between the model's output probability distribution and the absolute probability distribution, indicating the model's predictive alignment with empirical frequencies.
    \item \textbf{hd\_dist}: The pairwise cosine similarity between the hidden states of tokens across each layer is reported, reflecting the closeness of feature representations.
    \item \textbf{early\_exit\_jsd}: Inspired by the DoLA framework~\cite{dola:iclr23}, the model's output layer's MLP is repurposed on the hidden states of each layer to estimate the early exit probability distribution. The Jensen-Shannon (JS) divergence from the actual output probability distribution is computed, offering insights into the model's internal decision pathways.
    \item \textbf{hd\_abs\_cov}: The absolute covariance matrix of hidden states in each layer. This matrix captures the patterns of absolute co-variation in the hidden states' fluctuations within a layer, indicative of the interdependencies among features.
    \item \textbf{hd\_gram}: Adapting the concept from image processing techniques~\cite{gram:ijcai17}, the GRAM matrix of hidden states in each layer reveals the correlations and stylistic dimensions encoded in the model's representations.
\end{itemize}

We also devise supplementary metrics to capture the relationships between pairs of poems:

\begin{itemize}
    \item \textbf{out\_entropy\_fluc\_sim}: We employ the Dynamic Time Warping (DTW) algorithm to assess the similarity in fluctuation patterns between the entropy sequences of two poems' output probabilities.
    \item \textbf{emb\_sim}: We compute metrics such as Wasserstein Distance (WMD) and the Fréchet Distance (FD) based on the average-pooled embeddings from the output layer, providing a measure of closeness in the embedding space.
    \item \textbf{hd\_pca\_mse} \& \textbf{hd\_pca\_ssim}: Inspired by the structural similarity index measure~\footnote{\url{https://en.wikipedia.org/wiki/Structural_similarity_index_measure}}, we evaluate the similarity of principal component matrices obtained through Principal Component Analysis (PCA) on the sequence dimension of hidden states for two poems, utilizing metrics such as Mean Squared Error (MSE) and Structural Similarity Index (SSIM) on the normalized feature representations.
\end{itemize}

Furthermore, we determined the Gini coefficient as a reflection of the expressive diversity within each anthology. Additionally, we developed a scoring model trained on the \textbf{\textit{sft}} model with expert-annotated samples to appraise other poems. 

We list a portion of the computation results and made them available on GitHub~\footnote{\url{https://github.com/ByteDance-high-level-creation/literary\_appreciation}}.

Once computed, all metrics are stored in an object database in the form of arrays, indexed by the hash value derived from the content of each poem.

\subsection{Pattern Summarization}

For the aforementioned metrics, we will undertake a comprehensive statistical analysis, calculating not only the mean and variance but also the percentiles, specifically the 10th (pc10), 50th (pc50), and 90th (pc90) percentiles. This approach will be complemented by an examination of individual samples to discern any unique characteristics they may exhibit. In the case of multi-layer metrics, our investigative focus will be directed towards the initial, medial, and terminal layers, as they represent critical stages in the model's processing hierarchy.

Upon detecting any emerging patterns, we will endeavor to extrapolate and apply these insights consistently across all layers, seeking to uncover underlying principles that may govern the model's behavior.

Our analytical methodology is structured into three distinct yet interrelated stages: hypothesis formulation, inductive reasoning, and empirical verification. Throughout each phase, we meticulously scrutinize an extensive array of literary works to substantiate our evaluative conclusions with robust, evidence-based arguments. The integrity and reliability of our literary assessments are thereby ensured.

Further elucidation of our analytical process, including the rationale behind our chosen metrics and the implications of our findings, will be presented in the subsequent section.

\section{Discussion}\label{sec:discuss}

Upon thorough experimental analysis, we identify several patterns that exhibit a high degree of reliability. We will proceed to elucidate these patterns, dissecting them from multiple perspectives to provide a comprehensive understanding.

\subsection{Perplexity}

In the realm of ancient Chinese poetry, there prevails a prevalent aesthetic principle encapsulated by the adage \textit{eschewing the commonplace, embracing the innovative}. This principle underscores the desirability of originality in poetic composition and a deliberate departure from overused, clichéd expressions.

Perplexity, as a metric, indicates a model's predictive ability regarding the current text. Generally, within a single model, the perplexity of a text is inversely related to the model's ease of generation; that is, lower perplexity means greater ease of generation.

Consequently, we propose the hypothesis that a poetic work characterized by greater novelty or divergence from established linguistic conventions would likely exhibit higher perplexity, reflecting the model's relative unfamiliarity with such expressions. In contrast, works that are more canonical or have gained widespread circulation would be anticipated to present lower perplexity, owing to the model's increased exposure and, by extension, its enhanced predictive capabilities for such texts.

Here, we used the \textit{\textbf{base}} model to calculate perplexity, and the results are as shown in Tab~\ref{base_ppl:group}, Tab~\ref{base_ppl:author1} and Tab~\ref{base_ppl:author2}.

\begin{table*}
\centering
\begin{tabular}{p{0.6cm}p{1cm}p{1cm}p{1.2cm}p{1.2cm}p{1.2cm}p{0.8cm}p{0.8cm}p{1.4cm}p{1.0cm}p{0.8cm}}
\hline
    \textbf{base\newline \_ppl} & \textit{labelled \newline \_good} & \textit{labelled \newline \_normal} & \textit{historical \newline \_famous} & \textit{historical \newline \_normal} & ShiSanbai & JiuGe & Singer & TongGuang & JiangXi & XiKun \\
\hline
   avg & 8.07 & 7.66 & 7.23 & 7.45 & 5.36 & 7.04 & 6.85 & 8.51 & 7.02 & 6.62 \\
   std & 1.22 & 1.19 & 1.71 & 0.96 & 0.67 & 1.02 & 0.89 & 1.14 & 1.21 & 1.45 \\
\hline
\end{tabular}
\caption{\label{base_ppl:group}
    Perplexity of the \textbf{\textit{base}} model for different anthologies
}
\end{table*}

\begin{table*}
\centering
\begin{tabular}{p{0.6cm}p{0.6cm}p{1.2cm}p{0.6cm}p{0.6cm}p{1.2cm}p{1.0cm}p{0.8cm}p{0.6cm}p{0.8cm}p{1.2cm}p{0.6cm}p{1.2cm}}
\hline
    \textbf{base\newline  \_ppl} & Du \newline Fu & Li \newline Shangyin & Du \newline Mu & Bai \newline Juyi & Fan \newline Chengda & Chen \newline Shidao & Chen \newline Yuyi & Su \newline Shi & Yang \newline Wanli & Huang \newline Tingjian & Lu \newline You & Mei \newline Yaochen \\
\hline
    avg & 4.94 & 5.21 & 6.22 & 6.42 & 7.17 & 7.39 & 7.10 & 6.33 & 7.24 & 7.17 & 6.84 & 7.34 \\
    std & 1.82 & 1.50 & 1.07 & 0.90 & 0.92 & 0.85 & 0.84 & 1.03 & 0.84 & 1.05 & 0.74 & 0.88 \\
\hline
\end{tabular}
\caption{\label{base_ppl:author1}
    Perplexity of the \textbf{\textit{base}} model for different authors in the Tang and Song dynasties
}
\end{table*}

\begin{table*}
\centering
\begin{tabular}{p{0.6cm}p{1.0cm}p{1.2cm}p{0.6cm}p{1.2cm}p{1.2cm}p{1.2cm}p{1.0cm}}
\hline
    \textbf{base\newline \_ppl} & Qu \newline Dajun & Huang \newline Jingren & Li \newline E & Gong \newline Zizhen & Zhang \newline Wentao & Zheng \newline Zhen & Chen \newline Sanli \\
\hline
    avg & 8.03 & 7.87 & 7.99 & 8.40 & 8.01 & 8.92 & 8.72 \\
    std & 0.87 & 0.87 & 0.69 & 1.21 & 0.72 & 0.79 & 0.76 \\
\hline
\end{tabular}
\caption{\label{base_ppl:author2}
    Perplexity of the \textbf{\textit{base}} model for different authors in the Ming and Qing dynasties
}
\end{table*}

Our observations indicate that the perplexity values for the \textit{labelled\_good} category and poems of TongGuang are notably elevated. The distinctive linguistic features of TongGuang, often employing non-canonical expressions, corroborate our hypothesis. Despite JiuGe exhibiting a relatively high perplexity, meticulous review by experts disclosed the presence of lexical elements that diverge from the native speaker's linguistic intuition.

Furthermore, the variance in perplexity scores for the \textit{historical\_famous} subset was markedly higher than that observed for \textit{historical\_normal}. This discrepancy indirectly suggests a greater diversity in the literary styles among eminent authors. Works by illustrious figures of the early Tang and Song dynasties, such as Du Fu and Li Shangyin, have achieved broad dissemination and are probable inclusions in the \textbf{\textbf{base}} model's pre-training dataset. Their widespread familiarity to the model likely accounts for the remarkably low perplexity scores attributed to these texts.

\subsection{Entropy}

Information entropy serves as a pivotal measure for assessing the degree of dispersion or concentration within a distribution. In the context of a model's output probability distribution, an elevated entropy signifies a greater diversity of expressions, whereas a lower entropy suggests a more concentrated distribution. While perplexity encapsulates the overall unpredictability of textual elements, entropy delves into the granularity of local variations. Guided by this hypothesis, we undertook the computation of statistical values for the entropy sequences of all poetic works in the \textit{\textbf{base}} model. The ensuing findings are presented in Tab~\ref{base_ent:group}, Tab~\ref{base_ent:author1} and Tab~\ref{base_ent:author2}.

\begin{table*}
\centering
\begin{tabular}{p{1.2cm}p{1cm}p{1cm}p{1.2cm}p{1.2cm}p{1.2cm}p{0.8cm}p{0.8cm}p{1.4cm}p{1.0cm}p{0.8cm}}
\hline
    \textbf{base\newline \_entropy} & \textit{labelled \newline \_good} & \textit{labelled \newline \_normal} & \textit{historical \newline \_famous} & \textit{historical \newline \_normal} & ShiSanbai & JiuGe & Singer & TongGuang & JiangXi & XiKun \\
\hline
   avg & 5.50  & 5.40  & 5.27  & 5.19  & 4.69  & 4.99  & 5.16  & 5.74  & 5.34  & 5.05 \\
   std & 2.77  & 2.78  & 2.88  & 2.77  & 2.47  & 2.46  & 2.73  & 2.77  & 2.84  & 2.84 \\
\hline
\end{tabular}
\caption{\label{base_ent:group}
    Entropy of the \textbf{\textit{base}} model for different anthologies
}
\end{table*}

\begin{table*}
\centering
\begin{tabular}{p{1.2cm}p{0.6cm}p{1.2cm}p{0.6cm}p{0.6cm}p{1.2cm}p{1.0cm}p{0.8cm}p{0.6cm}p{0.8cm}p{1.2cm}p{0.6cm}p{1.2cm}}
\hline
    \textbf{base\newline  \_entropy} & Du \newline Fu & Li \newline Shangyin & Du \newline Mu & Bai \newline Juyi & Fan \newline Chengda & Chen \newline Shidao & Chen \newline Yuyi & Su \newline Shi & Yang \newline Wanli & Huang \newline Tingjian & Lu \newline You & Mei \newline Yaochen \\
\hline
    avg &  4.51 & 4.76 & 5.23 & 5.32 & 5.54 & 5.87 & 5.56 & 5.40 & 5.65 & 5.64 & 5.39 & 5.64 \\
    std &  3.04 & 2.88 & 2.65 & 2.68 & 2.66 & 2.65 & 2.71 & 2.68 & 2.62 & 2.69 & 2.63 & 2.60 \\
\hline
\end{tabular}
\caption{\label{base_ent:author1}
    Entropy of the \textbf{\textit{base}} model for different authors in the Tang and Song dynasties
}
\end{table*}

\begin{table*}
\centering
\begin{tabular}{p{1.2cm}p{1.0cm}p{1.2cm}p{0.6cm}p{1.2cm}p{1.2cm}p{1.2cm}p{1.0cm}}
\hline
    \textbf{base\newline \_entropy} & Qu \newline Dajun & Huang \newline Jingren & Li \newline E & Gong \newline Zizhen & Zhang \newline Wentao & Zheng \newline Zhen & Chen \newline Sanli \\
\hline
    avg & 5.76 & 5.66 & 5.65 & 6.38 & 5.78 & 6.39 & 6.04 \\
    std & 2.59 & 2.60 & 2.56 & 2.75 & 2.62 & 2.54 & 2.60 \\
\hline
\end{tabular}
\caption{\label{base_ent:author2}
    Entropy of the \textbf{\textit{base}} model for different authors in the Ming and Qing dynasties
}
\end{table*}

The analytical findings for both \textbf{base\_entropy} and \textbf{base\_ppl} exhibit a general alignment, albeit with subtle distinctions in their overall disparity. 

Notably, in the case of Qilv, the entropy trajectory from the initial to the final couplet typically delineates a pattern of initial decline followed by a subsequent ascent, often attaining its nadir at the second couplet. This observation, while not tabulated here, implies that the stylistic conventions of the second couplet may be more rigidly defined in comparison to its counterparts.

In the context of Ci, it is frequently observed that the entropy of the latter section is marginally elevated in comparison to the preceding section. This tendency may be construed as indicative of a richer thematic and content diversity in the concluding segment of these poetic compositions.

\begin{table}
\centering
\begin{tabular}{ccc}
\hline
    \textbf{base\_entropy} & \textit{Spring Outing} & \textit{Mourning} \\
\hline
    avg & 5.27 & 5.60 \\
    std & 2.61 & 2.76 \\
\hline
\end{tabular}
\caption{\label{base_ent:topic}
    Entropy of the \textbf{\textit{base}} model for different topics
}
\end{table}

In our thematic analysis, we deliberately chose works associated with the themes of \textit{Spring Outing} and \textit{Mourning}. Our investigation revealed that the entropy of poems pertaining to \textit{Mourning} markedly exceeds that of those associated with \textit{Spring Outing}. This observation intimates that within the canon of classical Chinese poetry, the articulation of sorrowful sentiments exhibits a richer tapestry of expressions. This diversity is poetically encapsulated by the adage: \textit{The landscapes may mourn, yet the poets find their muse}.

\begin{table*}
\centering
\begin{tabular}{cccc}
\hline
    \textbf{base\_entropy\_dtw} & \textit{inner\_labelled\_good} & \textit{inner\_labelled\_normal} & \textit{outer\_labelled\_good\_vs\_normal} \\
\hline
    avg & 100.80 & 98.87 & 102.55 \\
    std & 17.93 & 16.08 & 16.22 \\
\hline
\end{tabular}
\caption{\label{base_ent:dtw}
    DTW similarity of different entropy sequences of the \textbf{\textit{base}} model
}
\end{table*}

We further endeavored to quantify the correlation within entropy sequences. Utilizing a random sampling approach, we selected pairs of poems from both the \textit{labelled\_good} and \textit{labelled\_normal} subsets. Additionally, we compared pairs directly across these two subsets. 

For this analysis, we employed the Dynamic Time Warping (DTW) algorithm, a technique renowned for assessing the similarity in the geometric shapes of two temporal sequences.The data presented in Tab~\ref{base_ent:dtw} demonstrate a notable degree of similarity in the entropy sequence attributes across both high-caliber and average literary works. Nonetheless, these distinctions were observed to be rather subtle, offering only a broad reflection of the general trend rather than distinct patterns.

\subsection{Probability}

Another metric that captures the novelty and originality of a poetic composition is the absolute probability\footnote{For ease of reading, we convert all probabilities into percentages.} of its constituent tokens. In this paper, we assemble a world probability distribution~\footnote{\url{https://github.com/ByteDance-high-level-creation/literary_appreciation/blob/master/world_token_probs.json}} for all tokens, drawing from the BAAI-CCI 2.0 dataset~\footnote{\url{https://data.baai.ac.cn/details/BAAI-CCI2}}. Additionally, we procure and integrated probability distributions specific to ancient Chinese~\footnote{\url{https://github.com/ByteDance-high-level-creation/literary_appreciation/blob/master/ancient_token_probs.json}}, Qilv~\footnote{\url{https://github.com/ByteDance-high-level-creation/literary_appreciation/blob/master/qilv_token_probs.json}}, and Ci~\footnote{\url{https://github.com/ByteDance-high-level-creation/literary_appreciation/blob/master/ci_token_probs.json}}. The synthesis of these datasets has enabled us to derive an absolute probability table that is pertinent to the poetic genre.

The algorithmic procedure for deriving the absolute probability table is delineated as follows:

\lstset{frame=tb,
  language=Python,
  aboveskip=3mm,
  belowskip=3mm,
  % showstringspaces=false,
  columns=flexible,
  basicstyle={\small\ttfamily},
  numbers=none,
  breaklines=true,
  breakatwhitespace=true,
  tabsize=3
}
\begin{framed}
\begin{lstlisting}
block_size = 2048
token_cnt_arrs = np.zeros(tokenizer.vocab_size)

for file in file_list:
    for line in open(file, 'r'):
        for i in range(0, len(line), block_size):
            content_split = content[i:i+block_size]
            for token_id in tokenizer.encode(content_split):
                if token_id < tokenizer.vocab_size:
                    token_cnt_arr[token_id] += 1

np.save(file_to_save , token_cnt_arrs/token_cnt_arrs.sum())
\end{lstlisting}
\end{framed}

\begin{table*}
\centering
\begin{tabular}{p{1.0cm}p{1cm}p{1cm}p{1.2cm}p{1.2cm}p{1.2cm}p{0.8cm}p{0.8cm}p{1.4cm}p{1.0cm}p{0.8cm}}
\hline
    \textbf{abs\newline  \_prob} & \textit{labelled \newline \_good} & \textit{labelled \newline \_normal} & \textit{historical \newline \_famous} & \textit{historical \newline \_normal} & ShiSanbai & JiuGe & Singer & TongGuang & JiangXi & XiKun \\
\hline
   avg & 5.42 & 5.46 & 5.42 & 5.47 & 6.04 & 5.68 & 6.48 & 5.40 & 5.77 & 5.34 \\
   std & 6.40 & 6.71 & 6.38 & 6.52 & 6.98 & 6.38 & 7.09 & 6.25 & 6.60 & 6.14 \\
\hline
\end{tabular}
\caption{\label{abs_prob:group}
    Absolute probability of the \textbf{\textit{base}} model for different anthologies
}
\end{table*}

\begin{table*}
\centering
\begin{tabular}{p{1.0cm}p{0.6cm}p{1.2cm}p{0.6cm}p{0.6cm}p{1.2cm}p{1.0cm}p{0.8cm}p{0.6cm}p{0.8cm}p{1.2cm}p{0.6cm}p{1.2cm}}
\hline
    \textbf{abs\newline  \_prob} & Du \newline Fu & Li \newline Shangyin & Du \newline Mu & Bai \newline Juyi & Fan \newline Chengda & Chen \newline Shidao & Chen \newline Yuyi & Su \newline Shi & Yang \newline Wanli & Huang \newline Tingjian & Lu \newline You & Mei \newline Yaochen \\
\hline
    avg & 5.98 & 5.53 & 5.76 & 6.00 & 5.44 & 6.01 & 5.85 & 5.59 & 6.11 & 5.48 & 5.69 & 5.77 \\
    std & 6.92 & 6.20 & 6.40 & 7.00 & 6.32 & 6.68 & 6.65 & 6.48 & 6.86 & 6.36 & 6.50 & 6.73 \\
\hline
\end{tabular}
\caption{\label{abs_prob:author1}
    Absolute probability of the \textbf{\textit{base}} model for different authors in the Tang and Song dynasties
}
\end{table*}

\begin{table*}
\centering
\begin{tabular}{p{1.0cm}p{1.0cm}p{1.2cm}p{0.6cm}p{1.2cm}p{1.2cm}p{1.2cm}p{1.0cm}}
\hline
    \textbf{abs\newline  \_prob} & Qu \newline Dajun & Huang \newline Jingren & Li \newline E & Gong \newline Zizhen & Zhang \newline Wentao & Zheng \newline Zhen & Chen \newline Sanli \\
\hline
    avg & 6.26 & 5.65 & 5.79 & 5.32 & 5.79 & 5.27 & 5.18 \\
    std & 7.11 & 6.41 & 6.48 & 6.99 & 6.62 & 6.36 & 6.23 \\
\hline
\end{tabular}
\caption{\label{abs_prob:author2}
    Absolute probability of the \textbf{\textit{base}} model for different authors in the Ming and Qing dynasties
}
\end{table*}

Tab~\ref{abs_prob:group}, Tab~\ref{abs_prob:author1} and Tab~\ref{abs_prob:author2} present the absolute probability statistical metrics for different poetic anthologies. Our analysis discern that the absolute probabilities associated with historical poetic works are markedly lower compared to those of generated works. This phenomenon can be attributed to the propensity of LLMs to generate tokens with higher probabilities during the generative process. Nonetheless, it is essential to recognize that this metric, while indicative, does not provide an exhaustive reflection of the poetic quality. Tokens with higher probabilities are fully capable of culminating in the creation of exceptional poetic expressions.

\begin{table*}
\centering
\begin{tabular}{p{1.5cm}p{0.6cm}p{1.2cm}p{0.6cm}p{0.6cm}p{1.2cm}p{1.0cm}p{0.8cm}p{0.6cm}p{0.8cm}p{1.2cm}p{0.6cm}p{1.2cm}}
\hline
    \textbf{base\newline  \_prob\_kld} & Du \newline Fu & Li \newline Shangyin & Du \newline Mu & Bai \newline Juyi & Fan \newline Chengda & Chen \newline Shidao & Chen \newline Yuyi & Su \newline Shi & Yang \newline Wanli & Huang \newline Tingjian & Lu \newline You & Mei \newline Yaochen \\
\hline
    avg & 8.94 & 8.38 & 7.72 & 8.08 & 8.16 & 7.29 & 7.53 & 7.54 & 7.19 & 7.41 & 7.70 & 7.65 \\
    std & 8.81 & 8.34 & 7.97 & 9.48 & 9.97 & 8.68 & 8.70 & 8.05 & 8.27 & 8.26 & 8.38 & 8.87 \\
\hline
\end{tabular}
\caption{\label{prob_kld:author1}
    The KL divergence between output probability of the \textbf{\textit{base}} model and absolute probability for different authors in the Tang and Song dynasties
}
\end{table*}

\begin{table*}
\centering
\begin{tabular}{p{1.5cm}p{1.0cm}p{1.2cm}p{0.6cm}p{1.2cm}p{1.2cm}p{1.2cm}p{1.0cm}}
\hline
    \textbf{base\newline  \_prob\_kld} & Qu \newline Dajun & Huang \newline Jingren & Li \newline E & Gong \newline Zizhen & Zhang \newline Wentao & Zheng \newline Zhen & Chen \newline Sanli \\
\hline
    avg & 7.31 & 7.28 & 8.04 & 7.50 & 7.22 & 6.79 & 7.33 \\
    std & 8.67 & 8.28 & 9.84 & 10.47 & 8.39 & 8.63 & 9.08 \\
\hline
\end{tabular}
\caption{\label{prob_kld:author2}
    The KL divergence between output probability of the \textbf{\textit{base}} model and absolute probability for different authors in the Ming and Qing dynasties
}
\end{table*}

As Tab~\ref{prob_kld:author1} and Tab~\ref{prob_kld:author2} show, We proceeded to investigate the Kullback-Leibler (KL) divergence between the output probability distribution and the absolute probability distribution across the oeuvres of various authors. Our analysis unveiled an intriguing phenomenon: for authors such as Zheng Zhen, Huang Jingren, and Chen Shidao, whose writings are characterized by a lower absolute probability—employing a more distinctive lexical selection—the output conditional probabilities exhibit a convergence with the absolute probabilities. 

This finding underscores the unique creative approach of these authors, who, despite utilizing a less frequent vocabulary, manage to forge a close correspondence with the expected probabilities, thereby introducing a novel dimension to their poetic expressions. The discernible alignment between conditional and absolute probabilities in their works offers a profound insight into the distinctive poetic sensibilities that drive the generation of original and evocative literary works.

Regarding the early exit Jensen-Shannon Divergence (JSD) metric discussed in DoLA~\cite{dola:iclr23}, the original text posits that if a model's early exit in the final layers closely mirrors the final output, it signifies that the corresponding token is laden with substantial information, thereby warranting meticulous scrutiny. 

\begin{figure*}
    \centering
    \includegraphics[width=1.0\linewidth]{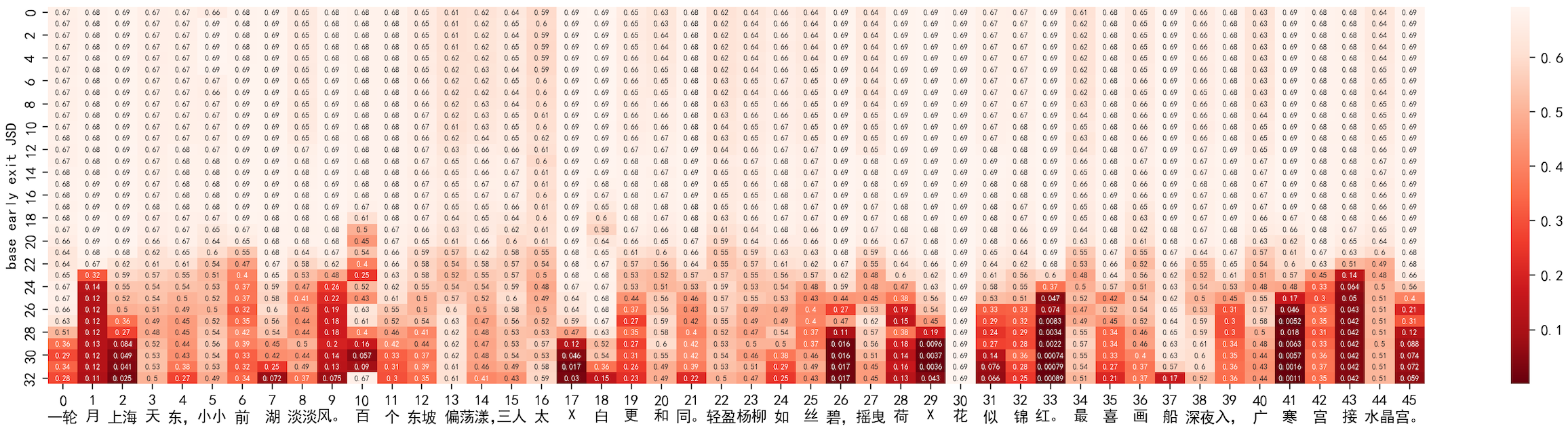}
    \caption{Visualization of the \textbf{\textit{base}} model early exit JSD in a poem of subpar caliber} 
    \label{fig:dola1_base}
\end{figure*}

\begin{figure*}
    \centering
    \includegraphics[width=1.0\linewidth]{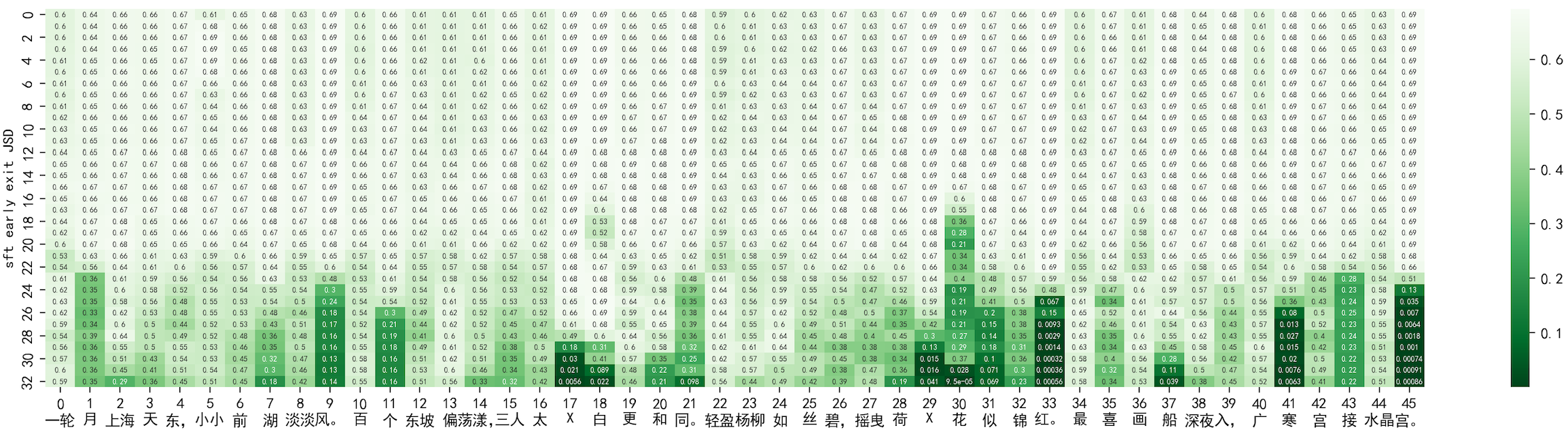}
    \caption{Visualization of the \textbf{\textit{sft}} model early exit JSD in a poem of subpar caliber} 
    \label{fig:dola1_sft}
\end{figure*}

\begin{figure*}
    \centering
    \includegraphics[width=1.0\linewidth]{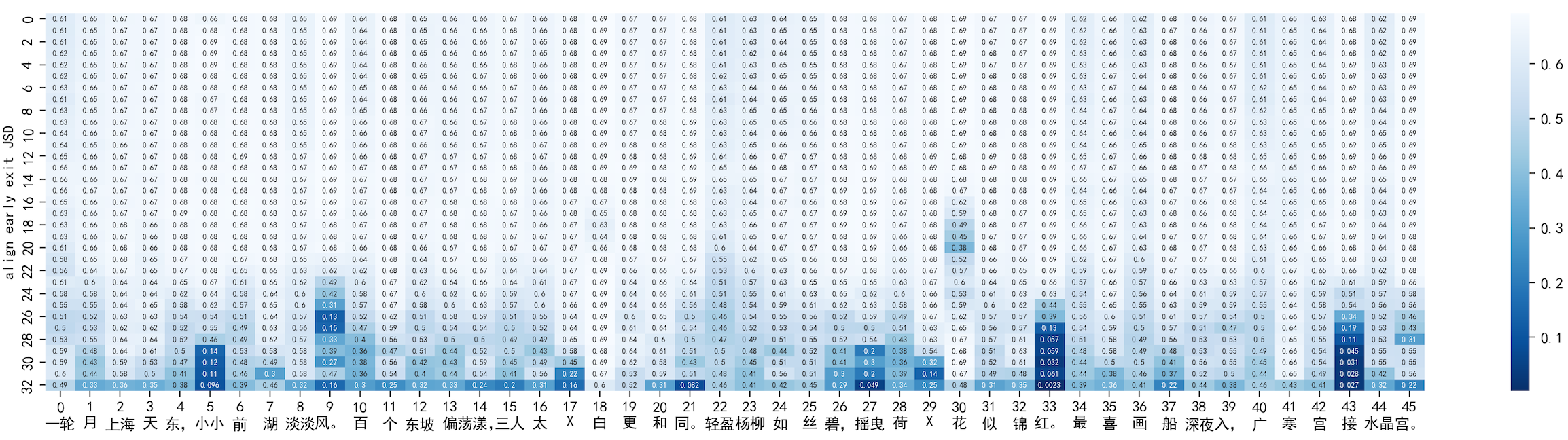}
    \caption{Visualization of the \textbf{\textit{align}} model early exit JSD in a poem of subpar caliber} 
    \label{fig:dola1_align}
\end{figure*}

\begin{figure*}
    \centering
    \includegraphics[width=1.0\linewidth]{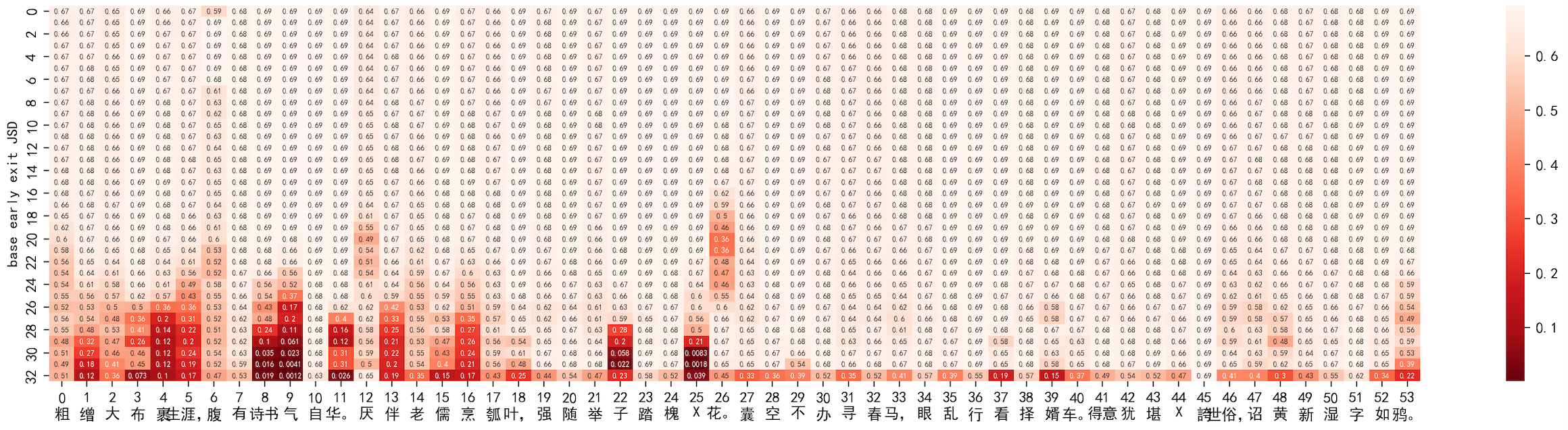}
    \caption{Visualization of the \textbf{\textit{base}} model early exit JSD in a poem of exceptional quality} 
    \label{fig:dola2_base}
\end{figure*}

\begin{figure*}
    \centering
    \includegraphics[width=1.0\linewidth]{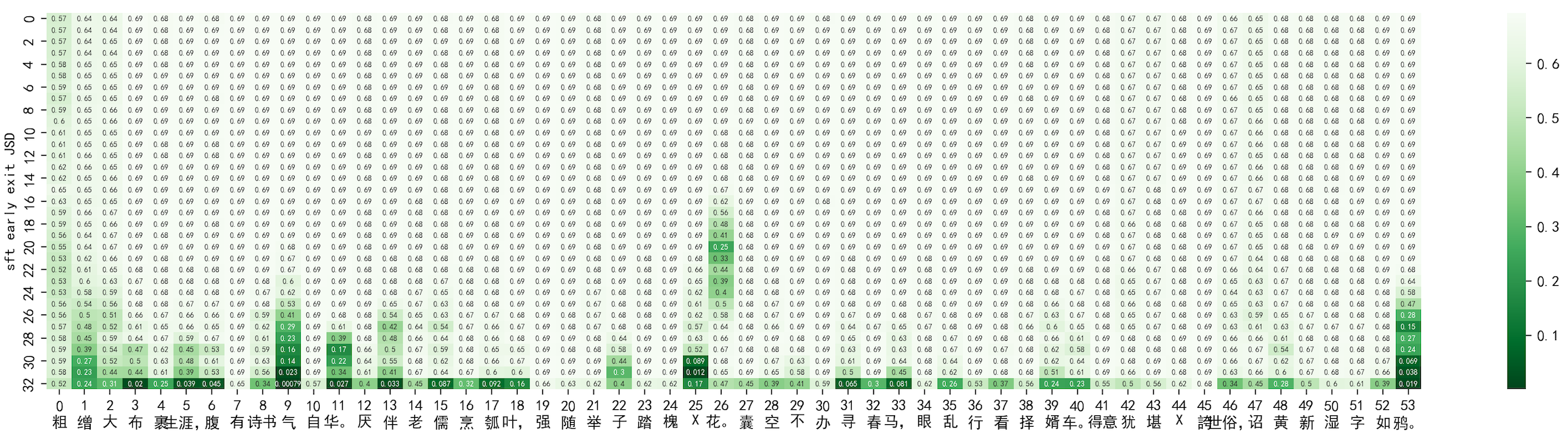}
    \caption{Visualization of the \textbf{\textit{sft}} model early exit JSD in a poem of exceptional quality} 
    \label{fig:dola2_sft}
\end{figure*}

\begin{figure*}
    \centering
    \includegraphics[width=1.0\linewidth]{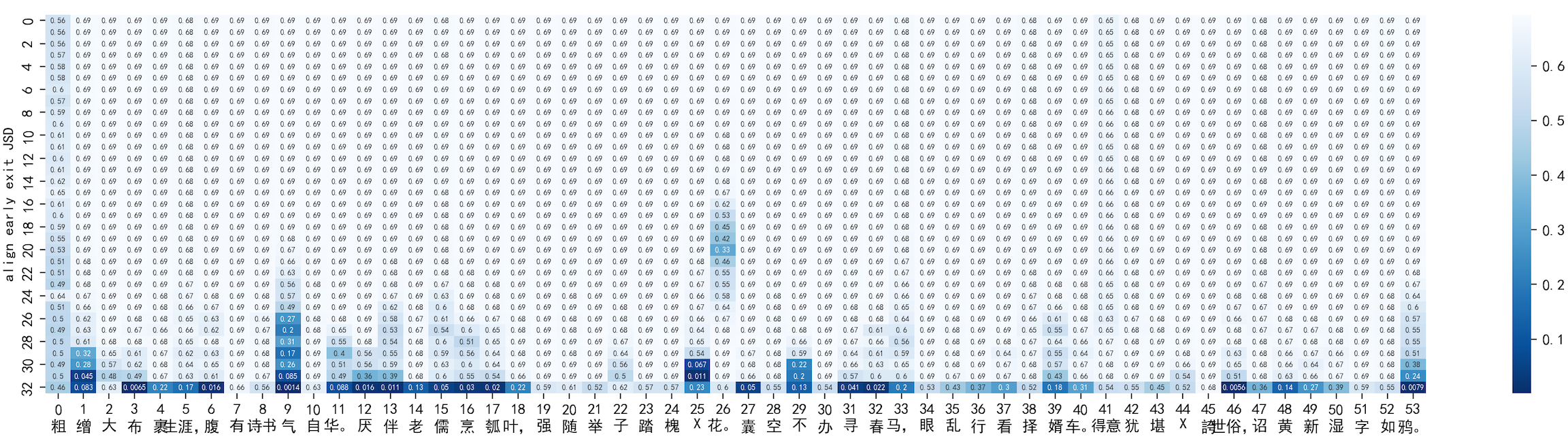}
    \caption{Visualization of the \textbf{\textit{align}} model early exit JSD in a poem of exceptional quality} 
    \label{fig:dola2_align}
\end{figure*}

However, our experimental findings suggest a contrasting phenomenon within the domain of classical poetry composition. As illustrated in Fig~\ref{fig:dola1_base}, Fig~\ref{fig:dola1_sft} and Fig~\ref{fig:dola1_align}, the poem penned by a Qing Dynasty bard, is notably rigid in its expression, bordering on the prosaic. In stark contrast, As shown in Fig~\ref{fig:dola2_base}, Fig~\ref{fig:dola2_sft} and Fig~\ref{fig:dola2_align}, the esteemed piece by Su Shi, is characterized by a markedly vivacious expression, demonstrating a vivid contrast in poetic vigor.

Hence, we propose the hypothesis that if the early exit predictions in the intermediate layers of the model closely resemble the final output, as illustrated in Fig~\ref{fig:dola1_base}, Fig~\ref{fig:dola1_sft} and Fig~\ref{fig:dola1_align}, this may suggest a propensity for the expression to lack originality. On the contrary, if the early exit predictions exhibit a higher degree of similarity to the final output layer as they approach the uppermost layers, as shown in Fig~\ref{fig:dola2_base}, Fig~\ref{fig:dola2_sft} and Fig~\ref{fig:dola2_align} it implies that the model has been engaged in a process of deep contemplation, deferring the final determination of wording until the latter stages. In these instances, the resulting expressions tend to exhibit a greater degree of innovation.

For more detailed analysis results of Qilv and Ci, please refer to here~\footnote{\url{https://github.com/ByteDance-high-level-creation/literary_appreciation/tree/master/Qilv}}~\footnote{\url{https://github.com/ByteDance-high-level-creation/literary_appreciation/tree/master/Ci}}.

\subsection{Embedding}

\begin{figure*}
    \centering
    \includegraphics[width=1.0\linewidth]{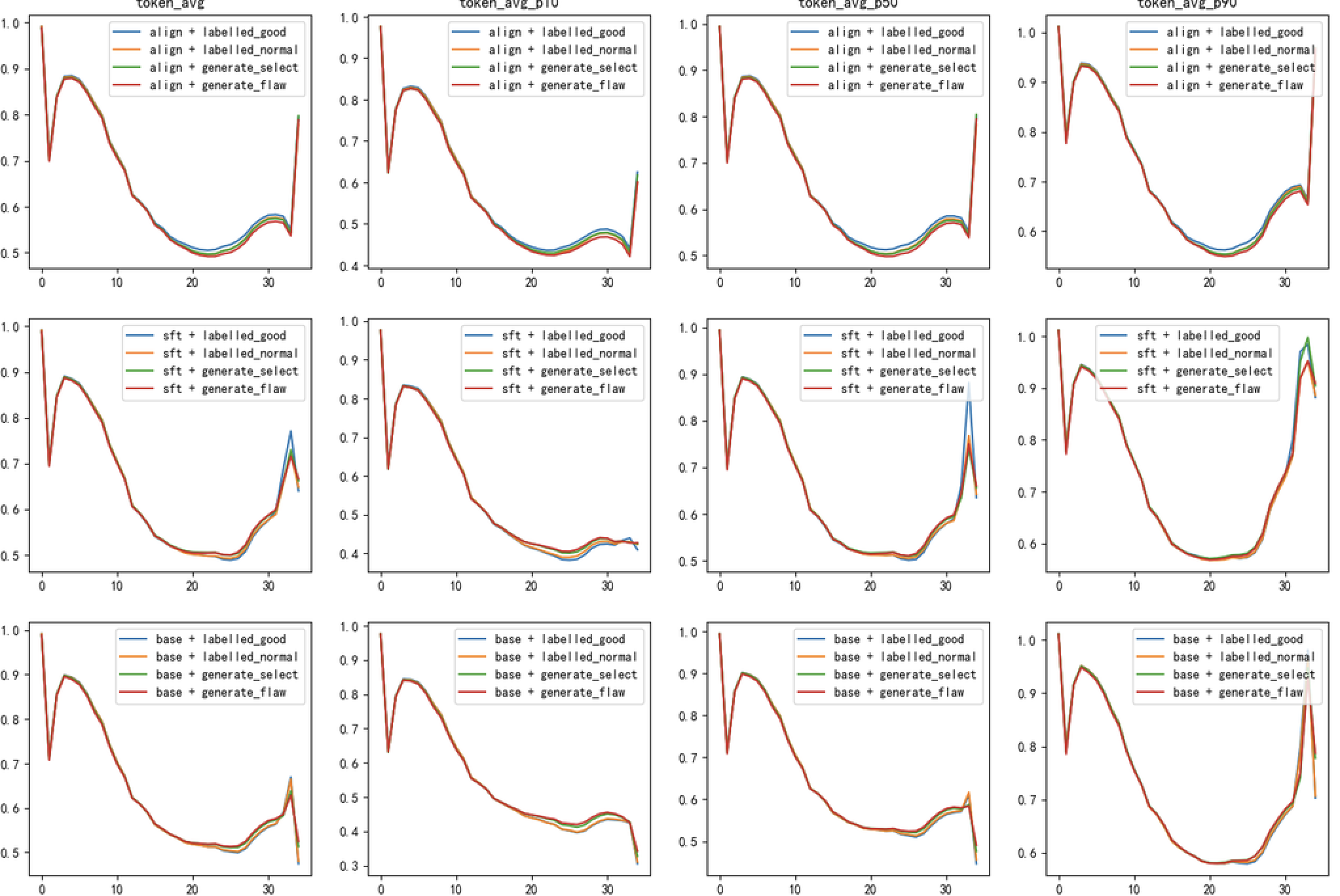}
    \caption{Visualization of the distances between the hidden states of each layer in Qilv} 
    \label{fig:emb_qilv}
\end{figure*}

\begin{figure*}
    \centering
    \includegraphics[width=1.0\linewidth]{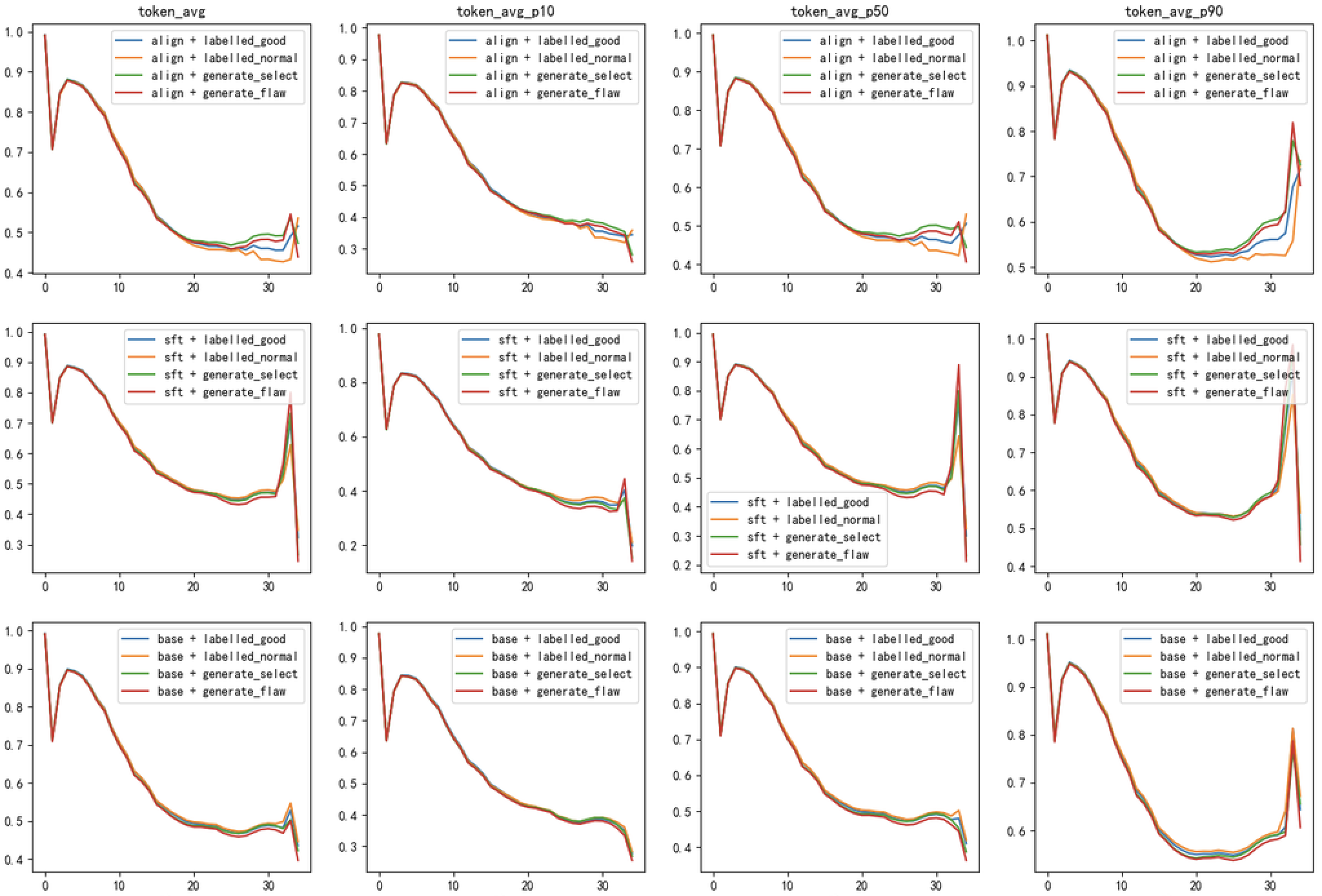}
    \caption{Visualization of the distances between the hidden states of each layer in Ci} 
    \label{fig:emb_ci}
\end{figure*}

We computed the pairwise distances between the hidden states of tokens across each layer of the model, utilizing the metric of 1 - cosine similarity, where a larger distance signifies greater dissimilarity. Fig~\ref{fig:emb_qilv} depicts the hidden state distances for Qilv, whereas Fig~\ref{fig:emb_ci} illustrates those for Ci.

The overarching trend we observed is as follows: the bottom layer exhibits the maximum distance, which is followed by an initial surge in the first hidden layer, then a steady decline, culminating in a precipitous drop at the final output layer. It is noteworthy to mention that the \textit{\textbf{align}} results for the regulated verse are excluded, as we hypothesize that they may be influenced by the selection bias of the annotated samples. These insights will be instrumental in assessing and verifying the potential biases present in the annotated sample set.

Our hypothesis posits that within the bottom layer, prior to the application of attention mechanisms, the inter-token correlation is minimal. As the model progresses through an increasing number of attention layers, the relationships among tokens intensify, reaching the pinnacle of interconnectivity at the output layer. However, the initial ascent and the subsequent descent in the correlation curve remain unexplained phenomena, necessitating further investigation.

Moreover, we observe that the distinctions between the \textbf{\textit{base}}, \textbf{\textit{sft}}, and \textbf{\textit{align}} models are negligible in the lower layers of the model architecture and only become pronounced in the vicinity of the output layer. This observation suggests that the fine-tuning process predominantly modifies the parameters in the proximity of the output layer, thereby influencing the model's predictive behavior.

\subsection{Frequency}

Gini coefficient of character frequency can intuitively reflect the diversity of a collection of works. Referring to Tab~\ref{freq:group}, Tab~\ref{freq:author1} and Tab~\ref{freq:author2} we find that Gini coefficients of historical collections are much smaller than those of generated works. This indirectly proves that the content generated by current LLMs is monotonous, and the issue of diversity urgently needs to be addressed.

\begin{table*}
\centering
\begin{tabular}{p{1.0cm}p{1cm}p{1cm}p{1.2cm}p{1.2cm}p{1.2cm}p{0.8cm}p{0.8cm}p{1.4cm}p{1.0cm}p{0.8cm}}
\hline
    \textbf{freq\newline  \_gini} & \textit{labelled \newline \_good} & \textit{labelled \newline \_normal} & \textit{historical \newline \_famous} & \textit{historical \newline \_normal} & ShiSanbai & JiuGe & Singer & TongGuang & JiangXi & XiKun \\
\hline
    & 0.46 & 0.44 & 0.44 & 0.44 & 0.57 & 0.54 & 0.53 & 0.46 & 0.46 & 0.46 \\
\hline
\end{tabular}
\caption{\label{freq:group}
    Gini coefficient for different anthologies
}
\end{table*}

\begin{table*}
\centering
\begin{tabular}{p{1.0cm}p{0.6cm}p{1.2cm}p{0.6cm}p{0.6cm}p{1.2cm}p{1.0cm}p{0.8cm}p{0.6cm}p{0.8cm}p{1.2cm}p{0.6cm}p{1.2cm}}
\hline
    \textbf{freq\newline  \_gini} & Du \newline Fu & Li \newline Shangyin & Du \newline Mu & Bai \newline Juyi & Fan \newline Chengda & Chen \newline Shidao & Chen \newline Yuyi & Su \newline Shi & Yang \newline Wanli & Huang \newline Tingjian & Lu \newline You & Mei \newline Yaochen \\
\hline
     & 0.461 & 0.461 & 0.457 & 0.467 & 0.466 & 0.463 & 0.457 & 0.462 & 0.464 & 0.463 & 0.461 & 0.458 \\
\hline
\end{tabular}
\caption{\label{freq:author1}
    Gini coefficient for different authors in the Tang and Song dynasties
}
\end{table*}

\begin{table*}
\centering
\begin{tabular}{p{1.0cm}p{1.0cm}p{1.2cm}p{0.6cm}p{1.2cm}p{1.2cm}p{1.2cm}p{1.0cm}}
\hline
    \textbf{freq\newline  \_gini} & Qu \newline Dajun & Huang \newline Jingren & Li \newline E & Gong \newline Zizhen & Zhang \newline Wentao & Zheng \newline Zhen & Chen \newline Sanli \\
\hline
     & 0.464 & 0.464 & 0.457 & 0.463 & 0.462 & 0.463 & 0.462 \\
\hline
\end{tabular}
\caption{\label{freq:author2}
    Gini coefficient for different authors in the Ming and Qing dynasties
}
\end{table*}

\subsection{Model Score}

We also trained an additional scoring model based on expert-annotated data. As shown in Tab~\ref{score:author1} and Tab~\ref{score:author2}, we listed the pc90 and std score for works by the following renowned authors (excluding Chen Sanli and Gong Zizhen due to sample issues). The pc90 reflects the author's level of artistic creation, while std reflects their consistency in creation. Due to the limited annotated samples, the generalization ability of our scoring model may be somewhat insufficient. Specifically, the works of Du Fu, due to their historical influence (many imitations), do not score particularly high, while the scoring model seems to overrate the works of Chen Yuyi and Huang Jingren.
 
\begin{table*}
\centering
\begin{tabular}{p{1.0cm}p{0.6cm}p{1.2cm}p{0.6cm}p{0.6cm}p{1.2cm}p{1.0cm}p{0.8cm}p{0.6cm}p{0.8cm}p{1.2cm}p{0.6cm}p{1.2cm}}
\hline
    \textbf{score} & Du \newline Fu & Li \newline Shangyin & Du \newline Mu & Bai \newline Juyi & Fan \newline Chengda & Chen \newline Shidao & Chen \newline Yuyi & Su \newline Shi & Yang \newline Wanli & Huang \newline Tingjian & Lu \newline You & Mei \newline Yaochen \\
\hline
    pc90 & 4.81 & 4.70 & 4.78 & 4.50 & 4.98 & 4.76 & 5.15 & 4.92 & 4.77 & 4.84 & 4.89 & 4.75 \\
    std & 0.37 & 0.40 & 0.34 & 0.38 & 0.38 & 0.39 & 0.42 & 0.45 & 0.46 & 0.40 & 0.41 & 0.45 \\
\hline
\end{tabular}
\caption{\label{score:author1}
    Model score for different authors in the Tang and Song dynasties
}
\end{table*}

\begin{table*}
\centering
\begin{tabular}{p{1.0cm}p{1.0cm}p{1.2cm}p{0.6cm}p{1.2cm}p{1.2cm}}
\hline
    \textbf{score} & Qu \newline Dajun & Huang \newline Jingren & Li \newline E & Zhang \newline Wentao & Zheng \newline Zhen \\
\hline
    pc90 & 4.72 & 5.09 & 5.01 & 4.97 & 4.94 \\
    std & 0.51 & 0.39 & 0.34 & 0.41 & 0.44 \\
\hline
\end{tabular}
\caption{\label{score:author2}
    Model score for different authors in the Ming and Qing dynasties
}
\end{table*}

\section{Future Work}

The knowledge reserve required for literary appreciation is immense, making it difficult for ordinary people to grasp its mysteries. However, LLMs provide us with an extremely convenient research platform, allowing us to stand on the shoulders of giants and enjoy literature alongside our predecessors. This paper only conducted a preliminary exploration in the field of ancient Chinese poetry. In the future, we will continue to conduct research in more areas such as jokes, scripts, and novels, hoping to derive more general conclusions.

Additionally, from an application perspective, the results of this paper will be widely used in the exploration process of AI poetry creation. This includes assisting in training, guiding the generation process, and intervening in the generated results. We believe that in the near future, AI-generated literary works will see significant improvements.

\section{Conclusion}

This paper introduces an approach grounded in LLMs to decipher the nuances of literary texts, with a particular emphasis on the ancient Chinese poetic corpus. Our investigation has unveiled distinct patterns that characterize LLMs' comprehension of literary compositions. We are convinced that these discerned patterns hold the potential to significantly streamline the path towards AI's capability to craft high-caliber literary creations.

\clearpage

\bibliographystyle{acl_natbib}
\bibliography{custom}

\end{document}